\documentclass[runningheads]{llncs}

\newcommand{\eg}{e.g.\@}

\newcommand{\etal}{et al.\@}

\usepackage{algorithm}
\usepackage{algorithmic}
\usepackage{tikz}
\usepackage{comment}
\usepackage{amsmath,amssymb} 
\usepackage{color}
\usepackage{graphicx}
\usepackage{enumitem}
\usepackage{booktabs}
\usepackage{bbding}
\usepackage{multirow}
\usepackage{makecell}
\usepackage{bbm}
\definecolor{citecolor}{RGB}{34, 139, 34}
\usepackage[pagebackref=true,breaklinks=true,colorlinks,bookmarks=false, citecolor=citecolor]{hyperref}
\usepackage[accsupp]{axessibility}  

\makeatletter
\newcommand{\printfnsymbol}[1]{%
  \textsuperscript{\@fnsymbol{#1}}%
}
\makeatother

\begin{document}
\pagestyle{headings}
\mainmatter
\def\ECCVSubNumber{3017}  

\title{Self-Filtering: A Noise-Aware Sample Selection for Label Noise with Confidence Penalization}

\titlerunning{Self-Filtering}

\author{Qi Wei \and
Haoliang Sun\thanks{Corresponding author} \and
Xiankai Lu \and
Yilong Yin\printfnsymbol{1}}
\authorrunning{Wei et al.}

\institute{School of Software, Shandong University, Jinan, China 
\email{\{1998v7,haolsun.cn,carrierlxk\}@gmail.com,  ylyin@sdu.edu.cn}\\}

\maketitle
\begin{abstract}
Sample selection is an effective strategy to mitigate the effect of label noise in robust learning. Typical strategies commonly apply the small-loss criterion to identify clean samples. However, those samples lying around the decision boundary with large losses usually entangle with noisy examples, which would be discarded with this criterion, leading to the heavy degeneration of the generalization performance. In this paper, we propose a novel selection strategy, \textbf{S}elf-\textbf{F}il\textbf{t}ering (SFT), that utilizes the fluctuation of noisy examples in historical predictions to filter them, which can avoid the selection bias of the small-loss criterion for the boundary examples. Specifically, we introduce a memory bank module that stores the historical predictions of each example and dynamically updates to support the selection for the subsequent learning iteration. Besides, to reduce the accumulated error of the sample selection bias of SFT, we devise a regularization term to penalize the confident output distribution. By increasing the weight of the misclassified categories with this term, the loss function is robust to label noise in mild conditions. We conduct extensive experiments on three benchmarks with variant noise types and achieve the new state-of-the-art. Ablation studies and further analysis verify the virtue of SFT for sample selection in robust learning.

\end{abstract}

\section{Introduction}
Neural networks exhibit notorious vulnerability to low-quality annotation. Especially, the generalization performance would heavily degrade when label noise arises. However, most existing datasets\cite{lu2020deep,lu2021segmenting} are commonly collected by Web crawlers, which inevitably contains label noise. Therefore, learning with noisy labels (LNL) poses great challenges for modern deep models~\cite{SUN2021108467}. 

Sample selection~\cite{wei2020combating,nguyen2019self,li2020dividemix,zhou2020robust,bai2021me} is an effective strategy to mitigate the effect of label noise in LNL. The main idea is to select the clean instances from the corrupted dataset by using a certain criterion and reduce the bias of the training set. 
Inspired by the \emph{memorization effect} that DNNs learn simple patterns shared by majority examples before fitting the noise~\cite{han2018co}, existing works~\cite{jiang2018mentornet,han2018co,yu2019does,wei2020combating,nguyen2019self,li2020dividemix,zhou2020robust,bai2021me} commonly apply the small-loss criterion that selects samples with loss value lower than a pre-defined threshold and treats them as clean instances. Although these methods exhibit favorable properties for LNL, selecting samples with small losses would discard those boundary examples, since they are with large losses and usually entangled with noisy instances. However, those boundary examples are essential for learning a good decision boundary. Moreover, the selection bias would be accumulated and further degenerate the generalization performance as the learning proceeds~\cite{han2018co,wei2020combating}. Besides, the loss threshold in this strategy is a crucial hyper-parameter that is usually carefully tuned by cross-validation, suffering from the issue of scalability and sensitivity.

\begin{figure}[t]
\centering
\begin{minipage}[t]{0.6\textwidth}
\centering
\includegraphics[width=7cm]{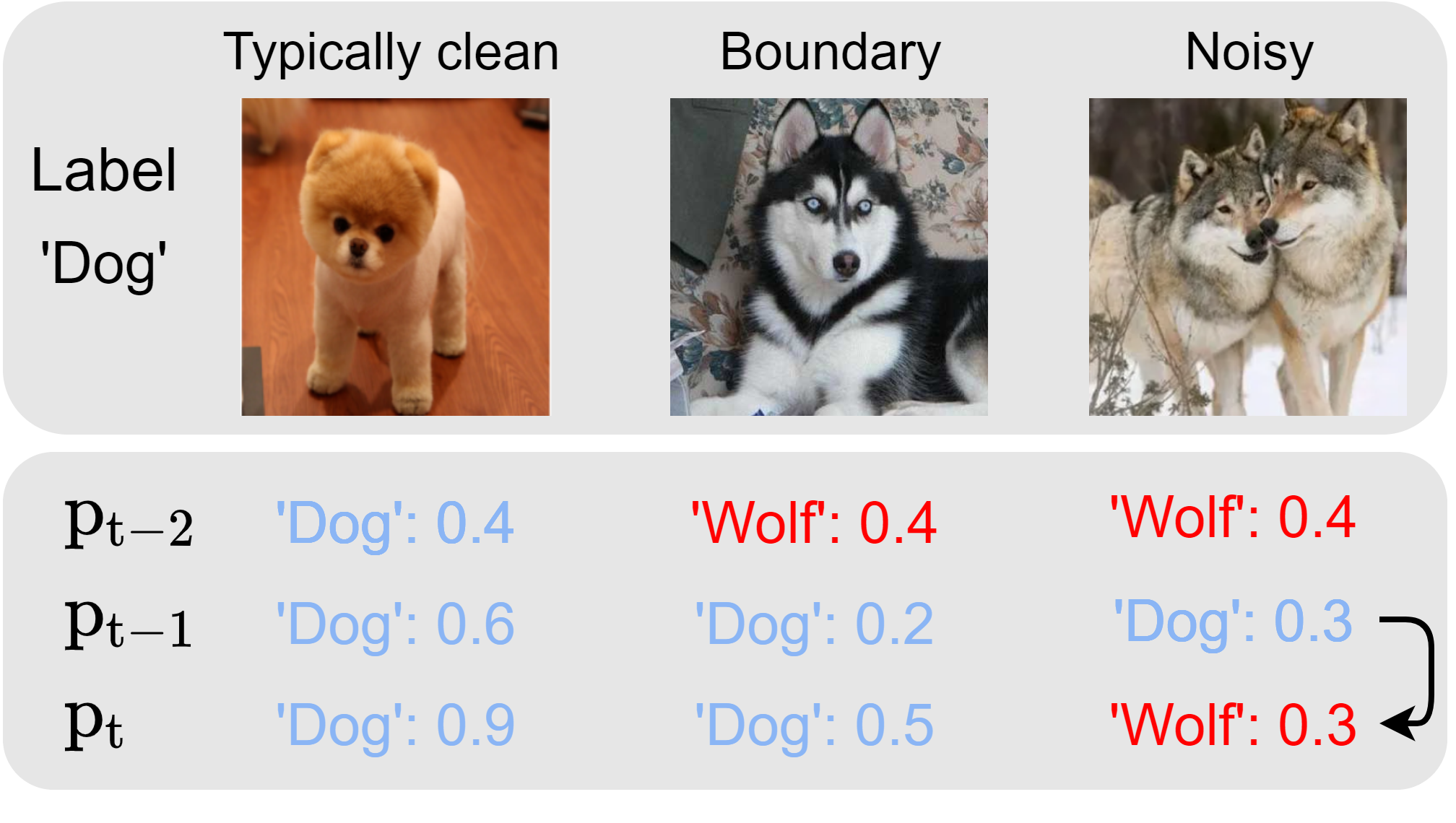}
\caption{Illustration of \textit{fluctuation} during model learning. Given a sample $(x,y)$, the fluctuation event is \textbf{correlated with predicted results and can be defined as the prediction $p_{t-1} = y$ at $t-1$ moment while $p_{t^*} \ne y \; (t^*>t-1)$ at subsequent moments}. The line with the arrow indicates the \textit{fluctuation} arises.}
\label{fig:introduction_fig}
\end{minipage}
\hspace{1mm}
\begin{minipage}[t]{0.34\textwidth}
\centering
\includegraphics[width=4.1cm]{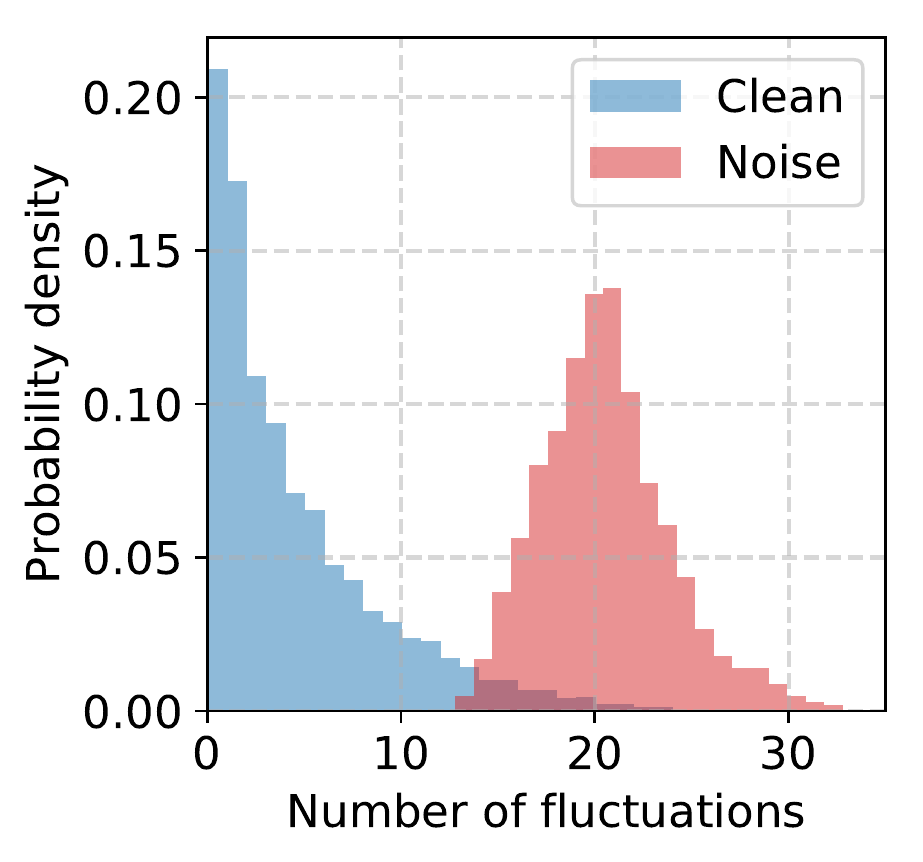}
\caption{Distributions of fluctuations number across synthetic CIFAR-10 with 40\% symmetric label noise after 100 epochs.}
\label{fig:1}
\end{minipage}
\end{figure}

In this paper, we have an interesting observation of the \textit{fluctuation} in LNL, which has the potential for identifying noisy samples and retaining boundary samples. As shown in Fig. \ref{fig:introduction_fig}, a fluctuation event occurs when \textit{a sample classified correctly at the current moment is misclassified in the following learning step}. Intuitively, as the learning proceeds, the discriminability of the classifier will be enhanced gradually, leading to a more accurate prediction for boundary examples rather than the noisy ones. Thus, the fluctuation would frequently occur for noisy samples but gradually disappear for boundary examples. On the other hand, more and more boundary samples are added to the training set via the fluctuation criterion, which can subsequently promote the learning of the classifier. To show the disparity between clean examples and noisy examples, we plot the distribution of the number of the fluctuation event in Fig. \ref{fig:1}. The clean and noisy sample can be essentially separated by the proposed fluctuation criterion.

Based on the above observation, we propose a novel sample selection strategy, Self-Filtering (SFT), that filters out noisy samples with the fluctuation in historical predictions. We store the historical predictions of each example for different training iterations in an external memory bank and detect the fluctuation event for each sample in the current moment. By applying this criterion, SFT can filter out noisy samples and retain boundary samples for current optimization. Meanwhile, the memory bank updates dynamically as the learning proceeds, which can support selecting more reliable samples to boost the model learning. To reduce the accumulated error stemming from the  selection bias of SFT, we design a regularization term to penalize the confident output distribution. By assigning the weight of the misclassified categories for the loss function, the model can avoid overconfidence of the correct prediction and further improve the robustness of our framework for severe label noise. We also integrate a semi-supervised method, FixMatch~\cite{sohn2020fixmatch}, into our framework to explore the useful information in the abandoned noisy samples, which has significantly advanced the model performance for noise-robust learning. 

The contribution can be summarized in four aspects. 
\begin{itemize}
\setlength{\itemsep}{0pt}
\setlength{\parsep}{0pt}
\setlength{\parskip}{0pt}
\item We rethink the sample selection in LNL and propose a new strategy to select clean samples by using the fluctuation of historical predictions.

\item We design a regularization term to penalise the confident output of the network, which can faithfully mitigate the sample selection bias.

\item We build a novel learning framework, Self-Filtering (SFT), and achieve the new state-of-the-art on three benchmarks with variant noise types.

\item We apply the proposed strategy to the prevailing learning framework and achieve a significant performance promotion, demonstrating its considerable versatility in LNL.
\end{itemize}


\section{Related Work}
\textbf{Sample selection.}
The majority of previous works exploit the memorization of DNNs and utilize the small-loss criterion to select clean samples~\cite{jiang2018mentornet,han2018co,yu2019does,wei2020combating}. One representative work is MentorNet~\cite{jiang2018mentornet} that proposes a teacher network to select clean samples for a student network with the small-loss trick. Similarly, Co-teaching~\cite{han2018co,yu2019does,wei2020combating} constructs a double branches network to select clean samples for each branch. A surrogate loss function~\cite{natarajan2013learning} is introduced to identify clean samples and theoretically guaranteed in~\cite{cheng2020learning}. To avoid tuning the threshold in the small-loss trick, Beta Mixture Model~\cite{arazo2019unsupervised} or Gaussian Mixture Model~\cite{li2020dividemix} is introduced to separate clean and noisy examples among the training loss automatically. Zhang \etal~\cite{zhang2020distilling} designs a meta-network trained with extra clean meta-data to identify noisy samples.  Recently, to achieve the stable prediction for sample selection, Model Ensemble (or Mean Teacher)~\cite{nguyen2019self} is introduced to compute the exponential moving-average predictions over past iterations and replace current predictions, which performs well to confront the more complex noise type (\emph{e.g.} instance-dependent label noise).
As a homologous approach that also aims to select the boundary samples, Me-Momentum ~\cite{bai2021me} modifies the training strategy and introduces two-loop training in curriculum learning.

\textbf{Robust loss function.}
A majority of robust loss functions have been theoretically analysed. Compared with the categorical cross-entropy (CCE) loss, mean absolute error (MAE) has been theoretically guaranteed to be robust to label noise~\cite{ghosh2017robust}. Based on this analysis, a novel generalized cross-entropy loss that combined CCE and MAE has been proposed in~\cite{zhang2018generalized} where its convergence and robustness are also analysed. Inspired by the symmetric Kullback-Leibler Divergence, a symmetric cross entropy (SCE)~\cite{wang2019symmetric} has been designed to mitigate the effect of noisy labels. To exploit the virtue of variant noise-robust losses, a meta-learning method is designed to learn to combine the four loss functions~\cite{amid2019robust,wang2019symmetric,gong2018decomposition,ghosh2017robust} adaptively.  
Recently, a new family of loss functions, the peer loss~\cite{liu2020peer} are proposed to punish the agreement between outputs of the network and noisy labels by adding a regularization term to cross-entropy. It has also been proved that minimizing those robust loss functions with corrupted labels is equivalent to minimise the cross entropy loss on the clean set under mild noise ratios.

\textbf{Label correction.}
The pioneering methods correct the noisy labels by an additional inference phase (\emph{e.g.} knowledge graphs~\cite{li2017learning} or graphical models~\cite{xiao2015learning}. Recently, two types of correction functions are proposed to correction. Firstly, transition matrix approaches~\cite{hendrycks2018using,xia2019anchor,yao2020dual} aim to construct a matrix that stores the flipping probability between the true label and the noisy one and is estimated with a set of clean data. Secondly, another family of methods utilize the output of the network to rectify labels. For example, Song \etal ~\cite{song2019selfie} proposes to select the clean samples by co-teaching framework~\cite{han2018co} and progressively refurbish noisy samples with the prediction confidence. To achieve learning the correction function in a data-driven way, Wu \etal~\cite{wu2020learning} builds a meta-correcter to generate labels with the input of the true label and previous predictions of meta-net.

Compared with Me-Momentum \cite{bai2021me} that selects samples in current epoch, we construct the memory bank and propose a novel criterion based on fluctuation by leveraging the historical predictions. Also, compared with existing works that select clean samples based on historical predictions (SELFIE \cite{song2019selfie} computes the entropy value from predictions histories and RoCL \cite{zhou2020robust} utilizes the variance of the training loss), our framework exhibits two advantages: (i) Less hyper-parameter. Our criterion contains only one hyper-parameter, namely, the size of historical predictions. However, the selection criteria in these works both contain another statistic threshold except the history size (\textit{e.g.}, an entropy threshold in \cite{song2019selfie} and a loss variance threshold in \cite{zhou2020robust}). (ii) Less sensitive. The setting of these thresholds is related to noise ratios, requiring more cross-validation processes. Those merits facilitate the application of our criterion to more general scenarios.

\section{Methods}
\subsection{Problem Definition and Overall}
Supposing that we have the training set $\mathcal{D} = \{ (\mathbf{x}^i, y^i)\}_{i=1}^n \in {(\mathcal{X},\mathcal{Y})}$ with corrupted labels and $y^i \in [{\rm K}]=\{1,2,...,k\}$, the output distribution of a classifier $f_\theta$ after $t$ training epochs can be written as $\mathbf{p}^{(t)} = f({\mathbf{x}};\theta)$. Here, $\theta$ is the learnable parameters. Learning with label noise (LNL) aims to find the optimal parameter $\theta^*$ which can achieve admirable generalization performance on the clean testing set.

For a majority of sample selection methods~\cite{han2018co,yu2019does,wei2020combating,nguyen2019self,li2020dividemix,zhou2020robust,bai2021me}, the robust training process for classifier $f$ can be summarized as the following phases. 
\par $\bullet$ Phase 1: select the reliable set $\mathcal{\tilde D}$ from the polluted dataset $\mathcal{D}$ via a certain selection strategy. (\emph{e.g.} small-loss criterion is designed to select the top $\tau\%$ of samples with the smaller loss values in the current mini-batch as the clean samples, where $\tau\%$ is the noise ratio estimated by cross-validation.)
\par $\bullet$ Phase 2: train the classifier $f$ on the selected set $\mathcal{\tilde D}$, and update the parameter as $\theta^{(t+1)} = \theta^t - \eta \nabla (\frac{1 } {{|\mathcal{\tilde D}|}} \sum\limits_{{(\mathbf{x},y) \in \mathcal{\tilde D}}} \mathcal{L}(\mathbf{x},y; \theta^t))$, where $\eta$ and $\mathcal{L}$ are the given learning rate and loss function, respectively.

\par $\bullet$ Phase 3: repeat the above phases until finding the optimal parameter $\theta^*$, then return the classifier $f$.

Our approach modifies the aforementioned two phases to render the network more robust on noisy labels. First, a novel criterion is proposed to select more boundary examples, which provides more decision information in the subset set $\mathcal{\tilde D}$. Second, we introduce a confidence regularization term to enable the loss function $\mathcal{L}$ more robust while tackling noisy labels. Eventually, we present the learning framework, Self-Filtering (SFT), for LNL that contains two stages of warming-up and main learning. To further exploit the useful knowledge in the discarded examples, we adopt the idea of semi-supervised learning and incorporate FixMatch into our SFT.

\subsection{Selecting with the fluctuation}
\begin{figure}[t]
    \centering  
    \includegraphics[width=12cm]{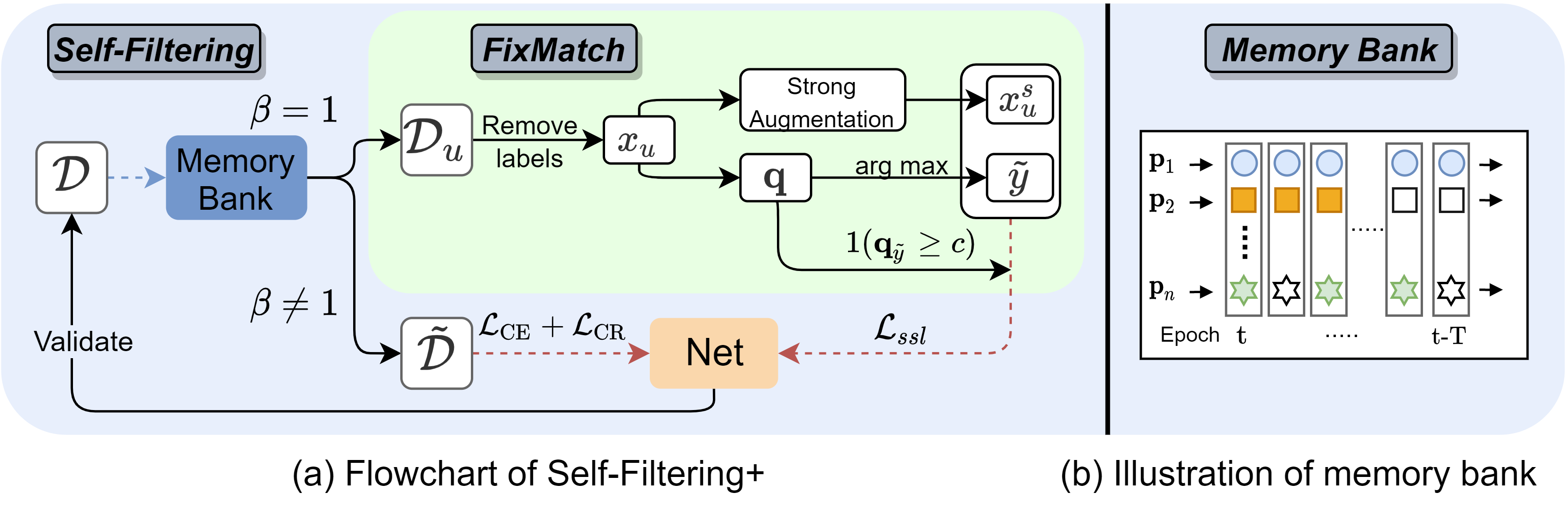}
    \caption{\textbf{(a)} The blue and red dashed line denote the update of memory bank and model parameter, respectively. For selected set $\mathcal{\tilde D}$, the objective function is $\mathcal{L}_{\rm CE}+\mathcal{L}_{\rm CR}$. For the set $\mathcal{ D}_u$, we remove their labels and input them to \emph{FixMatch} framework. \textbf{(b)} MB stores the past $T$ epoch prediction of each sample, and select the samples without fluctuation event to train the model. The updated classifier produces the prediction of the whole dataset to update MB.}
    \label{fig:flowchart}
\end{figure}

The key step in our selection strategy is to go through the historical prediction stored in the dynamically updated memory bank module. As shown in Fig.\ref{fig:flowchart} (b), we collect all predictions of the training set $\mathcal{D}$ for the epoch $t$ and store them in the memory bank $\mathcal{M}$. 
Specifically, this module contains those predictions of $T$ epochs in the memory bank, where the size of $\mathcal{M}$ is $n \times T$. $\mathcal{M}$ maintains a queen data structure with the principle of first-in-first-out (FIFO). Therefore, for epoch $t$, $\mathcal{M}$ stores predictions of the last $T$ epochs. Finally, for the example $(\mathbf{x},y)$ in the current epoch $t$, the criterion for identifying it as the fluctuated sample can be formulated as 
\begin{equation}\label{eq:beta}
    \beta = (\arg \max(\mathbf{p}^{t_1}) = y) \wedge (\arg \max(\mathbf{p}^{t_2}) \ne y),
\end{equation}
when $t_1,t_2 \in \{t-T,...,t\}$, $T \ge 2$ and $ t_1 < t_2$. A fluctuation event occurs when the sample classified correctly at the epoch $t_1$ is misclassified in the epoch $t_2$. By computing $\beta$ for each sample with $\mathcal{M}$, we discard these examples where $\beta=1$. Therefore, the clean samples selected by fluctuation criterion can be represented as:
\begin{equation}\label{eq:clean_d}
    \mathcal{\tilde{D}} = \{(\mathbf{x}^i,y^i)\in \mathcal{D} | \beta^i \ne 1 \}_{i=1}^n.
\end{equation}
The selected clean samples will be utilized in the following learning stage.

\subsection{Learning with selected examples}
SFT contains two stages: warming-up and main learning. We firstly conduct a warming-up of a few epochs to gain the basic discriminability for the network and achieve initialization for the memory bank module. Then, the main learning proceeds with clean samples selected via the fluctuation criterion. Also, the external memory bank module is dynamically updated for each epoch.

\noindent\textbf{Warming up}. It is necessary to warm up with the whole training set for the network before the main learning stage. However, it is usually vulnerable for pair noise that has a certain pattern for noise transition, especially under the extreme ratio (\textit{e.g.} 40\%). Therefore, we penalize the confidence for the output of the network to avoid radically moving forward in warm-up stage. Let $\mathbf{p}_y$ denotes the element in $\mathbf{p}$ for the label $y$, we formulate the confident-penalization regularization term as
\begin{equation}\label{eq:CR1}
\small
\begin{aligned}
\mathcal{R}  = -\alpha(\mathbf{p}_j) \cdot \log \mathbf{p}_j
\end{aligned}
\end{equation}
where $\mathbf{p}_j$ is the element with the second largest confidence of prediction $\mathbf{p}$.  Under the condition of the pair noise, we consider that the class $j$ could be the correct category for the noisy instance. $\alpha(\mathbf{p}_j)$ is an adaptive weight that can be computed by 
\begin{equation}\label{eq:alpha1}
\begin{aligned}
\alpha(\mathbf{p}_j) = \max (0, \Gamma - \frac{\mathbf{p}_j}{\mathbf{p}_y}).
\end{aligned}
\end{equation}
Here, $\Gamma$ is a hyper-parameter of the confidence threshold. If the network is over-confident in class $y$, it would be penalized with a larger $\alpha(\mathbf{p}_j)$. Finally, we can write our objective function of cross-entropy loss with the regularizer in the warming-up stage as 
\begin{equation}\label{eq:loss_wp}
\begin{aligned}
\mathcal{L} = \mathbb{E}_{(\mathbf{x}, y) \in \mathcal{D} }[ \mathcal{L}_{\rm CE}(f(\mathbf{x}, \theta), y)  +  \mathcal{R}(f(\mathbf{x}, \theta), y)] ,
\end{aligned}
\end{equation}

\noindent\textbf{Main learning.} By leveraging the proposed memory bank (MB) module, we can train the classification network with the selected clean examples. Meanwhile, the MB module is updated dynamically as the learning proceeds. Specifically, the clean sample is selected via the fluctuation criterion and subsequently utilized for training the network in the current epoch. After that, we collect the predictions for the whole examples and store them into the MB module for facilitating the selection in the next epoch. Since the classification network is weak in the early learning stage, the error would be accumulated in the following iteration step~\cite{han2018co}, leading to the selection bias. To tackle this problem, we propose a regularization term $\mathcal{L}_{\rm {CR}}$ to penalize the confident output distribution for an example $(\mathbf{x}, y)$ as following
\begin{equation}\label{eq:CR2}
    \mathcal{L}_{\rm {CR}} = - \frac{1}{\rm K} \sum\nolimits_{k \in [\rm K]}  {\alpha(\mathbf{p}_k)} \cdot \log \mathbf{p}_k,
\end{equation}
where the coefficient ${\alpha(\mathbf{p}_k)}$ can be computed by Eq. (\ref{eq:alpha1}).
The regularizer penalizes the confident output of the model by minimizing the expectation of the loss for each class. This is similar with the label smoothing (LS)~\cite{lukasik2020does} term. Recall LS of 
\begin{equation}\label{eq:ls}
\begin{aligned}
\mathcal{L}_{\rm LS}=-\log \mathbf{p}_{y} - \sum\nolimits_{k \in [\rm K]} \varepsilon \log \mathbf{p}_k,
\end{aligned}
\end{equation}
where $\varepsilon$ is a fixed smoothing coefficient and the later term can be regarded as a confidence regularization term. Compared with LS, our  coefficient $\alpha(\mathbf{p}_k)$ is adaptively computed by using the predictive value for each class, which can avoid tuning the hyper-parameter and can be more robust to the variant noise ratios.

Let $\mathcal{\tilde D}$ denotes the selected samples, the loss function for the main learning stage can be formulated as
\begin{equation}\label{eq:loss_function}
\begin{aligned}
\mathcal{L} =\mathbb{E}_{(\mathbf{x}, y) \in \mathcal{\tilde{D}}}[ \mathcal{L}_{\rm CE}(f(\mathbf{x}, \theta), y)  + \lambda  \mathcal{L}_{\rm {CR}}(f(\mathbf{x}, \theta), y) ],
\end{aligned}
\end{equation}
where $\lambda $ is a hyper-parameter set by cross-validation.

\subsection{Improving with FixMatch}
Our framework for sample selection is flexible, which can be combined with the state-of-the-art semi-supervised method. Hence, to further explore the knowledge in the discarded noise set, we introduce FixMatch~\cite{sohn2020fixmatch} to the main learning stage.
Since FixMatch is play-and-plug for SFT, we denote Self-Filtering with FixMatch as SFT+ in the following section. The SFT framework is flexible, which can be implemented by commonly-used differentiation tools. The whole learning framework is summarised in
Fig. \ref{fig:flowchart}(a). More details can be found in the supplemental material.

\section{Experiments}
To evaluate the performance of our proposed method, we implement experiments in several dimensions: (1) \textbf{Task variety}: we select three visual tasks with various dataset including  CIFAR-10~\cite{krizhevsky2009learning}, CIFAR-100~\cite{krizhevsky2009learning} and a real-world task Clothing1M~\cite{xiao2015learning}. (2) \textbf{Noise condition}: we manually corrupt the partial labels with three noise types (\emph{e.g.} symmetric, pair and instance dependent noise) on CIFAR-10\&100 and various noise ratios ranging from 20$\%$ to 80$\%$.   The code is available at \url{https://github.com/1998v7/Self-Filtering}.

\subsection{Noise types}
To simulate the actual noise condition in real-world, we refer to \cite{bai2021me} manually construct three noise types: symmetric, pair and instance-dependent label noise. Specially, we introduce a transition matrix $T$ to corrupt the clean label $y$ into a wrong label $\hat y$. Given a noise ratio $\tau$, for each sample $(\mathbf{x}, y)$,  the $T$ is defined as $T_{ij}(\mathbf{x})=\text{P}(\hat y = j|y = i)$, where $T_{ij}$ denotes that the true label transits from clean label $i$ to noisy label $j$.

(1) For \textbf{symmetric} label noise, the diagonal entries of the symmetric transition matrix are $1-\tau$ and the off-diagonal entries are $\tau / (k-1)$, where $k$ denotes the number of categories.

(2) For \textbf{pair-flipped} label noise, the diagonal entries of the symmetric transition matrix are $1-\tau$ and there exists other value $\tau$ in each row.

(3) For \textbf{instance-dependent} label noise, we stay the same construct algorithm with \cite{bai2021me}\cite{xia2020robust}. The actual flip rate relies on the pre-setting noise ratio $\tau$ and the representation of images. The detail algorithm is provided in Appendix 1.

(4) For \textbf{open-set} label noise, it's reported as the combination of aforementioned type noise. We select a real-world datasets to verify the effectiveness of our framework. Clothing1M\cite{xiao2015learning} contains one million images of 14 categories and its noise ratio is around 39.46\%.

\subsection{Network structure and experimental setup}
We adopt ResNet-18~\cite{he2016deep} and ResNet-34~\cite{he2016deep} to implement SFT on CIFAR-10 and CIFAR-100, respectively. The setting for the optimizer is listed as follows that SGD is with the momentum 0.9, the weight decay is 5e-4, the batch size is 32, and the initial learning rate is 0.02, and decayed with the factor 10 at 60 epoch. The number of epoch is set to be 75 for CIFAR-10 and 100 for CIFAR-100. For the warming-up stage, we train the network for 10 epochs and 30 epochs for CIFAR-10 and CIFAR-100, respectively, which is similar to Me-Momentum~\cite{bai2021me}. Typical data augmentations including randomly cropping and horizontally flipping are applied in our experiments.  
\begin{table*}[t]
\centering
\caption{Test accuracy (\%) on CIFAR-10 (the {\textit{top}}) and CIFAR-100 (the {\textit{bottom}}). The mean accuracy accuracy ($\pm$std) over 5 repetitions are reported.}
\scalebox{0.85}{
\begin{tabular}{lcccccc}
\toprule
\multicolumn{1}{c}{} &\multicolumn{2}{c}{Symm.} &\multicolumn{2}{c}{Pair.} &\multicolumn{2}{c}{Inst.} \\ \cline{2-7}
 Method     & 20\%             & 40\%             & 20\%              & 40\%              & 20\%              & 40\%\\\hline\hline
DMI~\cite{xu2019l_dmi}              & 88.18$\pm$0.36   & 83.98$\pm$0.48   & 89.44$\pm$0.41    & 84.37$\pm$0.78    & 89.14$\pm$0.36    & 84.78$\pm$1.97\\
Peer Loss~\cite{liu2020peer}        & 88.97$\pm$0.47   & 84.29$\pm$0.52   & 89.61$\pm$0.66    & 85.18$\pm$0.87    & 89.94$\pm$0.51    & 85.77$\pm$1.19\\
Co-teaching~\cite{han2018co}& 87.16$\pm$0.11   & 83.59$\pm$0.28   & 86.91$\pm$0.37    & 82.77$\pm$0.57    & 86.54$\pm$0.11    & 80.98$\pm$0.39\\
JoCoR~\cite{wei2020combating}       & 88.69$\pm$0.19   & 85.44$\pm$0.29   & 87.75$\pm$0.46    & 83.91$\pm$0.49    & 87.31$\pm$0.27    & 82.49$\pm$0.57\\
SELFIE~\cite{song2019selfie}    & 90.18$\pm$0.25   & 86.27$\pm$0.31   & 89.29$\pm$0.19    & 85.71$\pm$0.30      & 89.24$\pm$0.27    & 84.16$\pm$0.44 \\
CDR~\cite{xia2021robust} & 89.68$\pm$0.38 & 86.13$\pm$0.44 & 89.19$\pm$0.29 & 85.79$\pm$0.41    & 90.24$\pm$0.39 & 83.07$\pm$1.33\\
Me-Momentum~\cite{bai2021me}       & 91.44$\pm$0.33   & 88.39$\pm$0.34   & 90.91$\pm$0.45     & 87.49$\pm$0.56   & 90.86$\pm$0.21    & 86.66$\pm$0.91\\
PES\cite{bai2021understanding}   & 92.38$\pm$0.41  & 87.45$\pm$0.34 & 91.22$\pm$0.42  & 89.52$\pm$0.91   & \textbf{92.69$\pm$0.42}   & 89.73$\pm$0.51 \\ \hline
\textbf{SFT} {\small (ours)}      & \textbf{92.57$\pm$0.32}  & \textbf{89.54$\pm$0.27}    &\textbf{91.53$\pm$0.26} &\textbf{89.93$\pm$0.47}  &91.41$\pm$0.32  &\textbf{89.97$\pm$0.49}\\ \cline{1-7}
\hline\hline
DMI~\cite{xu2019l_dmi}              & 58.73$\pm$0.70   & 49.81$\pm$1.22   & 59.41$\pm$0.69    & 48.13$\pm$0.52    & 58.05$\pm$0.20    & 47.36$\pm$0.68\\
Peer Loss~\cite{liu2020peer}        & 58.41$\pm$0.55   & 50.53$\pm$1.31   & 58.73$\pm$0.51    & 50.17$\pm$0.42    & 58.91$\pm$0.41    & 48.61$\pm$0.78\\
Co-teaching~\cite{han2018co}& 59.28$\pm$0.47   & 51.60$\pm$0.49   & 58.07$\pm$0.61    & 49.79$\pm$0.69    & 57.24$\pm$0.69    & 49.39$\pm$0.99\\
JoCoR~\cite{wei2020combating}       & 64.17$\pm$0.19   & 55.97$\pm$0.46   & 60.42$\pm$0.35    & 50.97$\pm$0.58    & 61.98$\pm$0.39    & 50.59$\pm$0.71\\
SELFIE~\cite{song2019selfie}    & 67.19$\pm$0.30  & 61.29$\pm$0.39      & 65.18$\pm$0.23    & 58.67$\pm$0.51    & 65.44$\pm$0.43    & 53.91$\pm$0.66\\
CDR~\cite{xia2021robust}        & 66.52$\pm$0.24 & 60.18$\pm$0.22 & 66.12$\pm$0.31 & 59.49$\pm$0.47    & 67.06$\pm$0.50 & 56.86$\pm$0.62\\
Me-Momentum~\cite{bai2021me}        & 68.03$\pm$0.53   & 63.48$\pm$0.72   & 68.42$\pm$0.19    & 59.73$\pm$0.47    & 68.11$\pm$0.57    &58.38$\pm$1.28\\
PES~\cite{bai2021understanding}                 & 68.89$\pm$0.41  & 64.90$\pm$0.57   &  69.31$\pm$0.25  & 59.08$\pm$0.81  & 70.49$\pm$0.72  & 65.68$\pm$0.44 \\\hline
\textbf{SFT} {\small (ours)}      & \textbf{71.98$\pm$0.26}   & \textbf{69.72$\pm$0.31}   & \textbf{71.23$\pm$0.29 }    & \textbf{69.29$\pm$0.42}   & \textbf{71.83$\pm$0.42 }  & \textbf{69.91$\pm$0.54}\\
\bottomrule
\end{tabular}
}
\label{tab:cifar}
\end{table*}

\begin{table*}[t]
\centering
\caption{Comparison results with SSL with symmetric (\textbf{S}), pair (\textbf{P}) and instance (\textbf{I}) label noise.}
\scalebox{0.9}{
\begin{tabular}{cc|ccc|ccc}
\toprule
  &  & \multicolumn{3}{c|}{CIFAR-10} &  \multicolumn{3}{c}{CIFAR-100}   \\
Methods & SSL &\,\textbf{S} 50\%\, & \,\textbf{P} 40\%\, & \,\textbf{I} 40\%\,&\,\textbf{S} 50\%\, & \,\textbf{P} 40\%\, & \,\textbf{I} 40\%\,    \\\hline\hline
SELF~\cite{nguyen2019self} & {\scriptsize Mean Teacher}    & 91.4 & 90.9 & 90.4                                                        & 71.8              & 70.7              & 69.1  \\
CORES$^{2*}$~\cite{cheng2021learning} &{\scriptsize UDA}  & 93.1 & 92.4 & 92.2                                                        & 73.1              & 72.0              & 71.9  \\
DivideMix~\cite{li2020dividemix} &{\scriptsize MixMatch}  & 94.6              & \underline{93.4}  & 93.0              & 74.6              & 72.1              & 71.7              \\ 
ELR+ ~\cite{liu2020early} & {\scriptsize MixMatch}                             & 93.8              & 92.7              & 92.2              & 72.4              & \underline{74.4}  & 72.6              \\\hline
SFT+ & {\scriptsize FixMatch}                             & \underline{94.8}  & 92.9              & \textbf{94.4}     & \textbf{75.4}     & 74.2              & \underline{74.1}    \\ 
SFT+$^*$ & {\scriptsize MixMatch}                         & \textbf{94.9}     & \textbf{93.7}     & \underline{94.1}  & \underline{75.2}  & \textbf{74.9}     & \textbf{74.6}\\
 \bottomrule
\end{tabular}
}
\label{tab:ssl}
\end{table*}

For Clothing1M, we utilize the same architecture of ResNet-50 pre-trained on ImageNet. For image preprocessing and data augmentations, we resize the image to 256$\times$256 and crop them into 224$\times$224. The horizontally flipping is adopted. We train the classifier network for 15 epochs using SGD with 0.9 momentum, weight decay of 0.0005, and the batch size of 32. The warming-up stage is one epoch. The learning rate is set as 0.02 and decayed with the factor of 10 after 10 epochs. Following the convention from ~\cite{li2020dividemix}, we sample 1000 mini-batches from the training data while ensuring the labels (noisy) are balanced.

\noindent\textbf{Hyper-parameter setup.} We set memory bank size $T=3$ and confidence threshold $\Gamma = 0.2$ for all experiments. We set the trade-off coefficient $\lambda$ in loss function as 1 and the threshold $c$ in FixMatch as $0.95$ following ~\cite{sohn2020fixmatch}.

\begin{figure}[t]
{
\flushleft
\makeatletter\def\@captype{table}\makeatother
\begin{minipage}{0.35\textwidth}
\centering
		\caption{\small Test accuracy (\%) on the Clothing1M.}
		\scalebox{0.85}{
		\begin{tabular}{c|c|c}
        \toprule
        \multicolumn{2}{c|}{\scriptsize Method} & \scriptsize Acc.            \\ \hline\hline
        &\scriptsize Cross Entropy                         &  \scriptsize64.54        \\
        &\scriptsize MentorNet~\cite{jiang2018mentornet}  & \scriptsize 67.14           \\
        &\scriptsize Co-teaching~\cite{han2018co}         & \scriptsize 68.51         \\
   \scriptsize No  &\scriptsize JoCoR~\cite{wei2020combating}       & \scriptsize70.30   \\
    \scriptsize SSL &\scriptsize Forward~\cite{patrini2017making}     & \scriptsize69.84 \\
        &\scriptsize Joint Optim~\cite{tanaka2018joint} & \scriptsize72.23    \\
        &\scriptsize Me-Momentum~\cite{bai2021me}       & \scriptsize73.13    \\ 
        &\scriptsize\textbf{SFT}          & \scriptsize \textbf{74.16} \\ \hline
        &\scriptsize DivideMix~\cite{li2020dividemix}    & \scriptsize74.76  \\
    \scriptsize SSL    &\scriptsize ELR+~\cite{liu2020early}            &\scriptsize  74.81   \\
        &\scriptsize\textbf{SFT+*}     & \scriptsize\textbf{75.08} \\
        \bottomrule
        \end{tabular}
		 }
		\label{tab:clothin1m}
\end{minipage}
\hspace{1mm}
\makeatletter\def\@captype{figure}\makeatother
\begin{minipage}{0.62\textwidth}
    \centering
	\scalebox{0.9}{
        \includegraphics[width=8.3cm]{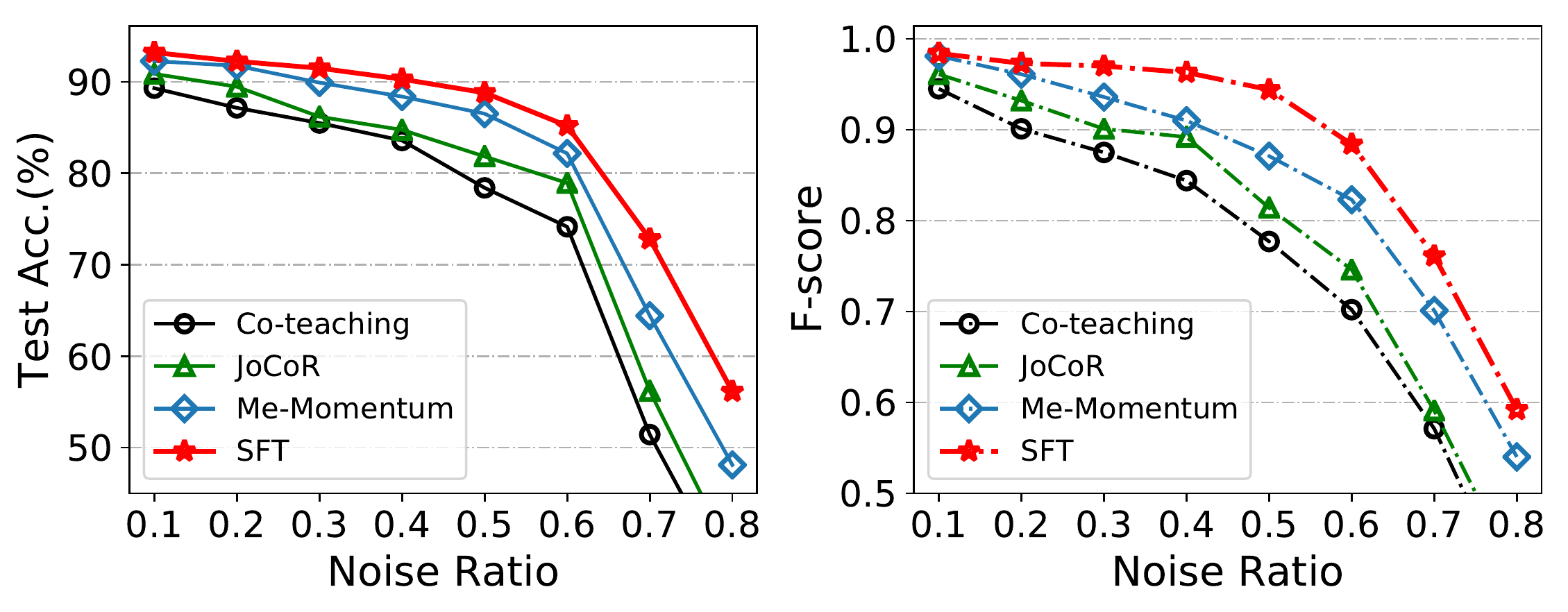} 
		}
		\caption{\small Illustration of the performance of variant methods as the noise ratio changes. Notably, SFT can produce considerable performance under extreme noise ratios.}
		\label{fig:extreme_nr}
\end{minipage}
}
\end{figure}

\begin{figure*}[t]
\centering
\includegraphics[width=12cm]{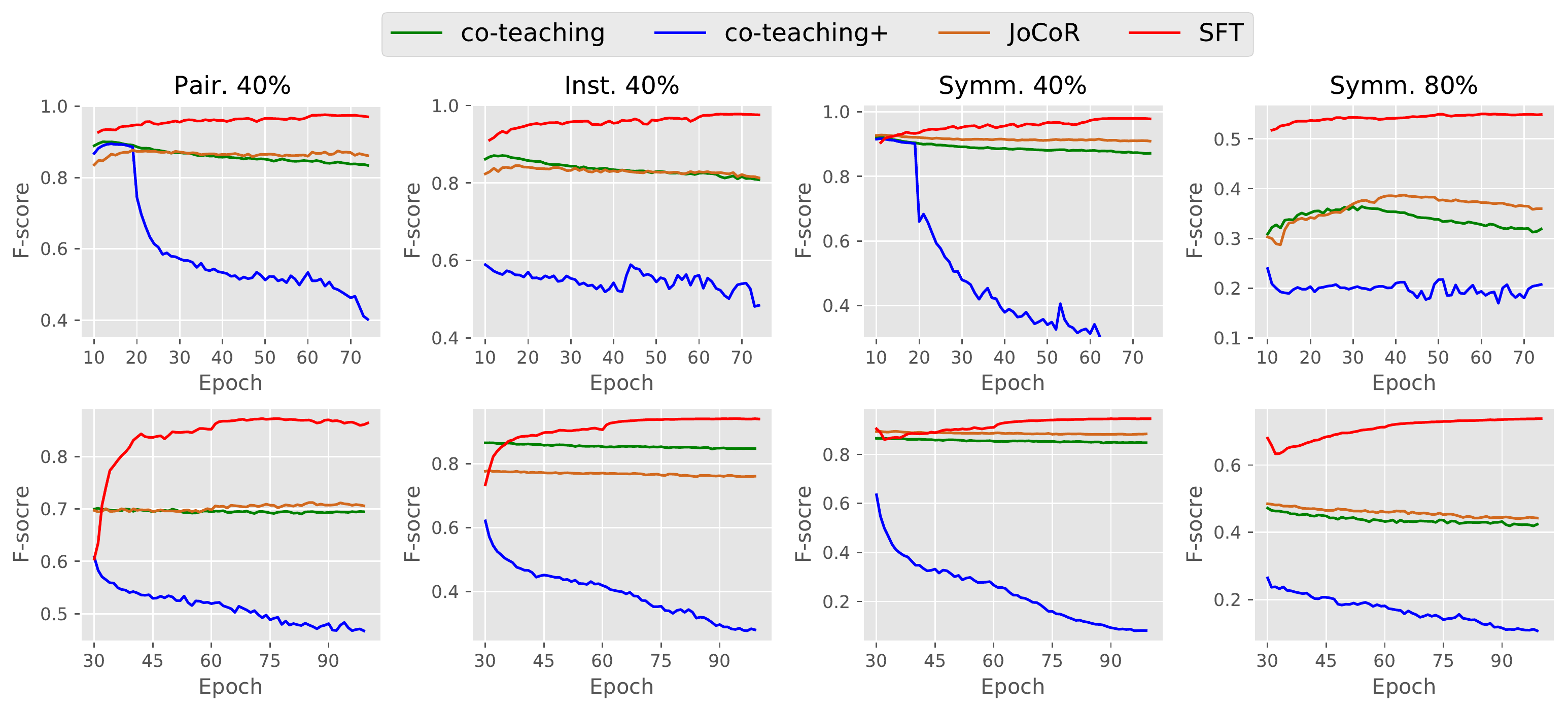}
\caption{\small SFT achieves the highest F-score for sample selection results on CIFAR-10 (the \textit{top}) and CIFAR-100 (the \textit{bottom}).}
\label{fig:acc of selected-samples}
\end{figure*}

\subsection{Comparison with state-of-the-arts}
\noindent\textbf{Baseline.} We evaluate our method against the following state-of-the-art methods.  (1) {Robust loss function}: DMI~\cite{xu2019l_dmi}, Peer loss~\cite{liu2020peer}.  (2) {Sample selection methods}: Co-teaching~\cite{han2018co}, JoCoR~\cite{wei2020combating}, SELFIE~\cite{song2019selfie},  Me-Momentum~\cite{bai2021me}. Note that both of them explore the memorization effect and utilize the small-loss criterion to select clean examples. (3) {Sample selection methods with SSL}: SELF~\cite{nguyen2019self}, CORES$^{2*}$~\cite{cheng2021learning}, DivideMix~\cite{li2020dividemix} and ELR+ ~\cite{liu2020early}. Specially, SELF and CORES$^{2*}$ use the SSL methods of Mean Teacher~\cite{tarvainen2017mean} and UDA~\cite{XieDHL020}, respectively. DivideMix and ELR+ both utilize MixMatch. (4) Others: CDR~\cite{xia2021robust}, PES~\cite{bai2021understanding}.

\noindent\textbf{Results on CIFAR-10 \& 100.} To evaluate the performance of SFT, we conduct experiments on CIFAR-10 and CIFAR-100 under three noise types with variant noise ratios $\tau \in \{0.2, 0.4\}$. Since SSL can dramatically improve the performance, we split the table of each benchmark into two parts for a fair comparison. The bottom part includes methods using SSL, while the top part does not. As shown in Tab. \ref{tab:cifar}, our SFT outperforms the almost state-of-the-art on two datasets with three noise types.  
Compared with the homologous approach Me-Momentum, SFT achieves the higher accuracy of 89.97\% and gains the significant improvement of ${\mathbf{6.24}}$\% in \emph{Symm.}-40\%, $\mathbf{9.56}$\% in \emph{Pair.}-40\% and $\mathbf{11.53}$\% in \emph{Inst.}-40\%. The results exhibit the superiority of SFT in handling the noise issue for learning with more categories. 

Our framework is flexible that can easily integrate the semi-supervised technology (SSL) to further boost the generalization performance, denoting SFT+ (with FixMatch) or SFT+$^*$ (with MixMatch). Tab. \ref{tab:ssl} shows the result when combining vanilla models with recent SSL techniques. The hybrid approach, SFT+ (or SFT+$^*$) consistently outperforms other methods. Especially, for the instance-dependent label noise, our approach achieves the average of $\mathbf{1.7}$\% improvement compared with the state-of-the-art.

\noindent\textbf{Results on Clothing1M.} To demonstrate the effectiveness of our method on real-world noisy labels, we evaluate our approach on Clothing1M dataset, which is a real-world benchmark in LNL tasks. As shown in Tab. \ref{tab:clothin1m}, our proposal obtains the state-of-the-art performance. For a fair comparison, we divide the table into two parts according to using the semi-supervised technique or not. Our SFT+* (with MixMatch) outperforms other methods and achieves the improvement of $\mathbf{0.27}\%$ over ELR+, demonstrating its effectiveness for the real-world application.

\begin{figure}[t]
\centering
\begin{minipage}[t]{0.48\textwidth}
\centering
 \includegraphics[width=5.8cm]{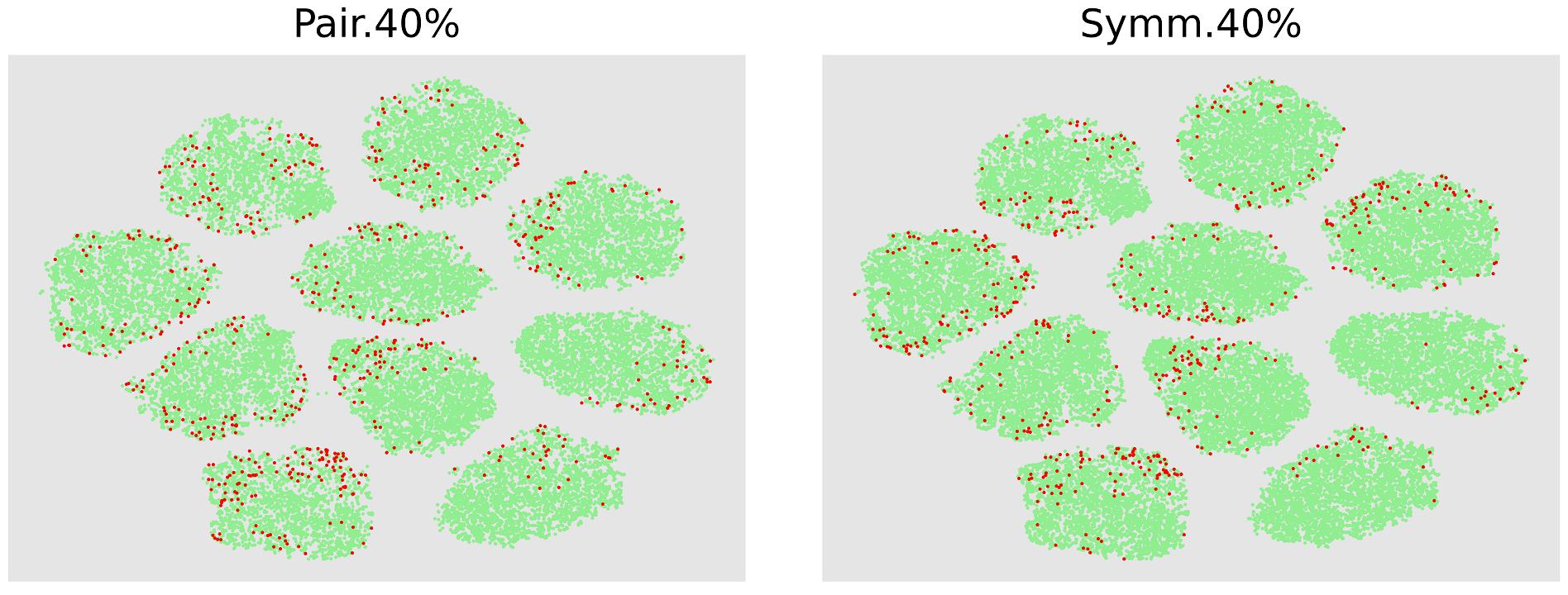}
\caption{\small Most boundary examples can be selected by SFT. The clusters with green dots are samples in the same categories. The red dots are the selected clean samples in the last ten epochs.}
\label{fig:cluster}
\end{minipage}
\hspace{1mm}
\begin{minipage}[t]{0.48\textwidth}
\centering
\includegraphics[width=6cm]{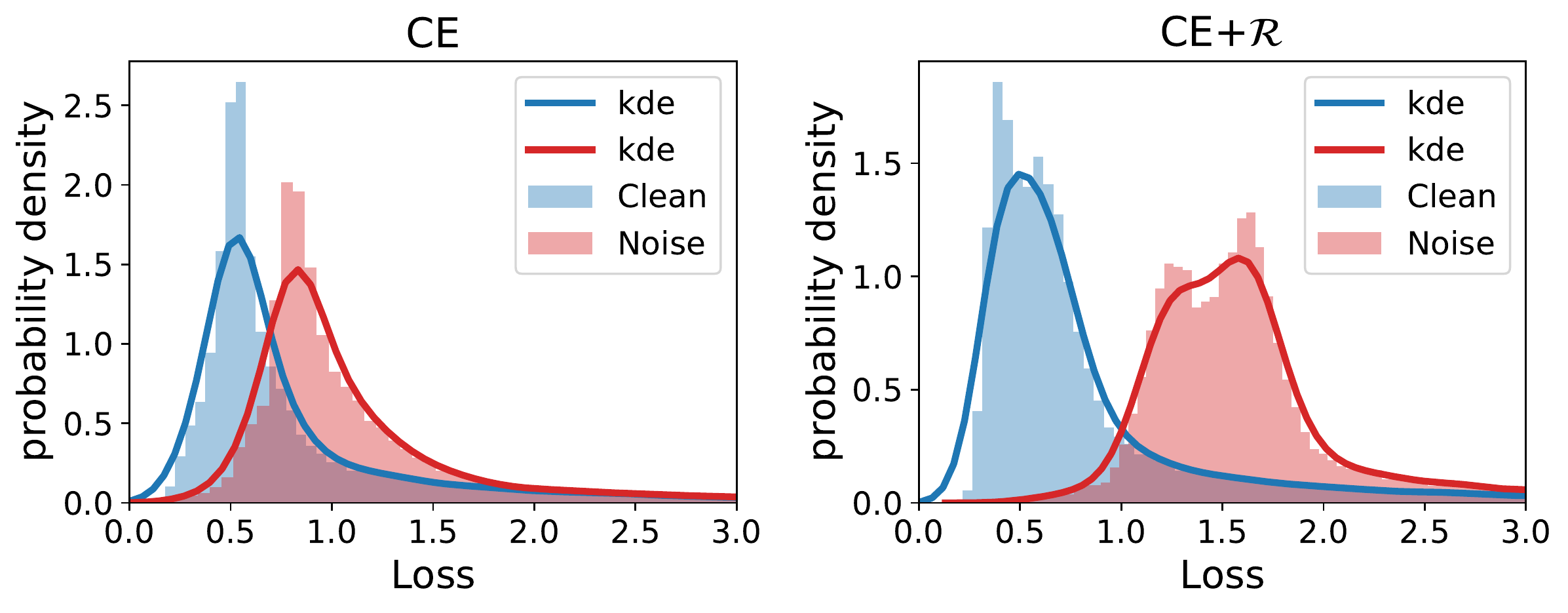} 
\caption{\small The losses distribution of noisy CIFAR-10 with Pair.40\% after warming-up. There exists an obvious disparity between clean and noisy samples by using the reguliazer $\mathcal{R}$ during warming-up.}
\label{fig:loss_distribution}
\end{minipage}
\end{figure}

\subsection{Further analysis}

\noindent\textbf{Robustness.} To validate the robustness of SFT in a more challenging noisy environment, we compare it with  three robust methods~\cite{han2018co,wei2020combating,bai2021me} under a more noise ratio setting $\tau \in \{0.1,...,0.8\}$.  We plot the test accuracy and F-score as the \emph{Symm.} noise ratio increases. As shown in Fig. \ref{fig:extreme_nr}, SFT produces a good performance, even on the challenging condition with a far higher noise ratio. Compared with the homologous approach Me-Momentum, SFT exhibits the favorable property of selecting the clean samples and consistently outperforms Me-Momentum on variant noise ratios. SFT also exhibits considerable robustness under extreme noise conditions as shown in Fig. \ref{fig:extreme_nr}.

\noindent\textbf{Effectiveness.} We evaluate the effectiveness of SFT from these three aspects. 

(1) \textit{How accurate is the sample selection strategy}? We conduct the comparison experiments on CIFAR-10 \& 100 and plot the curves of F-score for the selection result. As shown in Fig. \ref{fig:acc of selected-samples}, SFT achieves the highest F-score on all noise types as the training proceeds. Under 40\% noise ratios, SFT attains average 0.97 F1-score, indicating it obtains high selection accuracy and recall scores. 

(2) \textit{Are the boundary examples selected}? We conduct the experiment with \emph{Inst.} \&  \emph{Pair}-40\% and record the selected set in different epoch to illustrate the dynamic selecting process by t-SNE~\cite{van2008visualizing}. As illustrated in Fig. \ref{fig:cluster}, most of selected samples (red) lie around the decision boundary, demonstrating the effectiveness of the fluctuation criterion for selecting boundary examples.

(3) \textit{How effective is the framework in the warming-up stage}? We plot the distribution of training losses for all instances after warming-up in Fig. \ref{fig:loss_distribution}. The blue and red parts represent the losses of clean and noise labels, respectively. By introducing the regularizer during the warming-up stage, there exists an obvious disparity between clean and noisy samples, verifying the effect of $\mathcal{R}$ in mitigating the overconfidence during warming-up. 

\begin{table*}[t]
\small
\centering
\caption{\small The significant gain with our selection criterion for other methods.}
\scalebox{0.8}{
\begin{tabular}{cc|cccc|cccc}
\toprule
\multicolumn{2}{c|}{} &\multicolumn{4}{c|}{CIFAR-10}  &\multicolumn{4}{c}{CIFAR-100} \\
\multicolumn{2}{c|}{Method}   
& \textbf{P} 40\% & \textbf{I} 40\% & \textbf{S} 40\% & \textbf{S} 80\%        
& \textbf{P} 40\% & \textbf{I} 40\% & \textbf{S} 40\% & \textbf{S} 80\%\\ \hline\hline

Co-teaching & Base  & 82.7 & 80.8 & 83.5 & 34.6 
                    & 49.8 & 49.3 & 51.6 & 19.2\\
            & After 
            & 84.9\textcolor[RGB]{21, 152, 56}{(2.2)} \,
            & 83.9\textcolor[RGB]{21, 152, 56}{(3.1)} \,
            & 85.2\textcolor[RGB]{21, 152, 56}{(1.7)} \,
            & 39.7\textcolor[RGB]{21, 152, 56}{(5.1)} \,
            & 52.1\textcolor[RGB]{21, 152, 56}{(2.3)} \,
            & 51.9\textcolor[RGB]{21, 152, 56}{(2.6)} \,
            & 54.1\textcolor[RGB]{21, 152, 56}{(2.5)} \,
            & 23.6\textcolor[RGB]{21, 152, 56}{(4.4)}\\ \hline

JoCoR   & Base  & 83.9 & 82.4 & 85.4  & 38.0 
                & 51.0 & 50.6 & 56.0  & 22.7 \\
        & After 
        & 86.8\textcolor[RGB]{21, 152, 56}{(2.9)} \,
        & 85.6\textcolor[RGB]{21, 152, 56}{(3.2)} \,
        & 87.1\textcolor[RGB]{21, 152, 56}{(1.7)} \,
        & 41.6\textcolor[RGB]{21, 152, 56}{(3.6)} \,
        & 53.6\textcolor[RGB]{21, 152, 56}{(2.6)} \,
        & 53.6\textcolor[RGB]{21, 152, 56}{(3.0)} \,
        & 58.4\textcolor[RGB]{21, 152, 56}{(2.4)} \,
        & 26.8\textcolor[RGB]{21, 152, 56}{(4.1)}\\\hline
    
DivideMix & Base & 93.4 & 94.5 & 94.7  & 93.0 
                & 71.0 & 70.9 & 72.9  & 57.9 \\
        & After 
        & 94.1\textcolor[RGB]{21, 152, 56}{(0.7)} 
        & 94.7\textcolor[RGB]{21, 152, 56}{(0.2)} 
        & 94.6\textcolor[RGB]{255,0,   0}{(0.1)} 
        & 92.8\textcolor[RGB]{255,0,   0}{(0.2)} 
        & 72.4\textcolor[RGB]{21, 152, 56}{(1.4)}
        & 72.0\textcolor[RGB]{21, 152, 56}{(1.1)}
        & 73.3\textcolor[RGB]{21, 152, 56}{(0.4)}
        & 59.1\textcolor[RGB]{21, 152, 56}{(1.2)}\\
\bottomrule
\end{tabular}}
\label{tab:replacing}
\end{table*}

\begin{table}[t]
\centering
\caption{\small Comparison of the total training time (hours) on CIFAR-10.}
\scalebox{0.8}{
\begin{tabular}{cccccc|cc}
\toprule
 Co-teaching\cite{han2018co}\, & MW-Net\cite{shu2019meta}\, & MSLC\cite{wu2020learning}\, & CORES$^2$\cite{cheng2021learning}\, & DivideMix\cite{li2020dividemix}\,  & Me-Momentum\cite{bai2021me}\, & \, SFT \quad & SFT+\\ \hline
 2.9h         & 4.7h  & 4.1h & 2.6h       & 4.4h      & 4.8h    & \textbf{2.4h}  & 4.1h  \\
\bottomrule
\end{tabular}
}
\label{tab:time}
\end{table}

\noindent\textbf{Versatility.} Our fluctuation criterion is flexible and play-and-plug that can be applied on other modern methods~\cite{han2018co,wei2020combating,li2020dividemix}. We replace the sample selection phase with our selection module and conduct experiments on CIFAR-10 \& 100 under four settings of noise. As shown in Tab. \ref{tab:replacing}, by introducing the fluctuation criterion, these three methods almost outperform the basic version that uses the small-loss criterion. Even in the current SOTA work DivideMix, the improvement of performance on CIFAR-100 is remarkable. The fluctuation criterion gains a significant average improvement of almost $\mathbf{2.0}$\% under all settings. These results demonstrate the great flexibility of our proposed selection strategy.

\noindent\textbf{Efficiency.} 
We compare the training time with typical methods to show its efficiency. We evaluate them on CIFAR-10 and obtain the mean value of training time with the 40\% rate of three noise types. All models are trained on a single Geforce-3090. As shown in Tab.\ref{tab:time}, SFT is consistently faster than other methods since it can directly back-propagate with selected and does not rely on sophisticated learning strategies, \textit{\eg}, two-loop training in Me-Momentum~\cite{bai2021me}.

\begin{figure}[t]
{
\flushleft
\makeatletter\def\@captype{table}\makeatother
\begin{minipage}{0.44\textwidth}
\centering
	\caption{\small The hyper-parameter selection of $\Gamma$. Test acc. is reported.}
    \scalebox{1}{
    \begin{tabular}{c|cccc}
    \toprule
\scriptsize \textbf{CIFAR-10}   & \scriptsize 0.2        & \scriptsize 0.4       & \scriptsize 0.6       & \scriptsize 0.8 \\\hline
\scriptsize Symm. 40\% & \scriptsize \textbf{89.67}      & \scriptsize 89.51     & \scriptsize 89.27     & \scriptsize 89.29    \\
\scriptsize Pair. 40\% &  \scriptsize 89.74     & \scriptsize \textbf{90.11}     & \scriptsize 89.51     & \scriptsize 89.43     \\
\scriptsize Inst. 40\% & \scriptsize \textbf{89.93}      & \scriptsize 89.61     & \scriptsize 89.37     & \scriptsize 89.29    \\ \hline\hline
\scriptsize \textbf{CIFAR-100}  & \scriptsize 0.2        & \scriptsize 0.4  & \scriptsize 0.6  & \scriptsize 0.8 \\\hline
\scriptsize Symm. 40\% & \scriptsize \textbf{69.88}      & \scriptsize 69.61     & \scriptsize 69.17     & \scriptsize 69.34     \\
\scriptsize Pair. 40\% & \scriptsize \textbf{69.23}      & \scriptsize 69.07     & \scriptsize 68.71     & \scriptsize 68.66      \\
\scriptsize Inst. 40\% & \scriptsize 69.84      & \scriptsize \textbf{69.91}     & \scriptsize 69.52     & \scriptsize 69.37 \\ 
    \bottomrule
\end{tabular}
}
\label{tab:ablation_gamma}
\end{minipage}
\hspace{0.5mm}
\makeatletter\def\@captype{figure}\makeatother
\begin{minipage}{0.52\textwidth}
    \centering
		\scalebox{0.8}{
        \includegraphics[width=7.8cm]{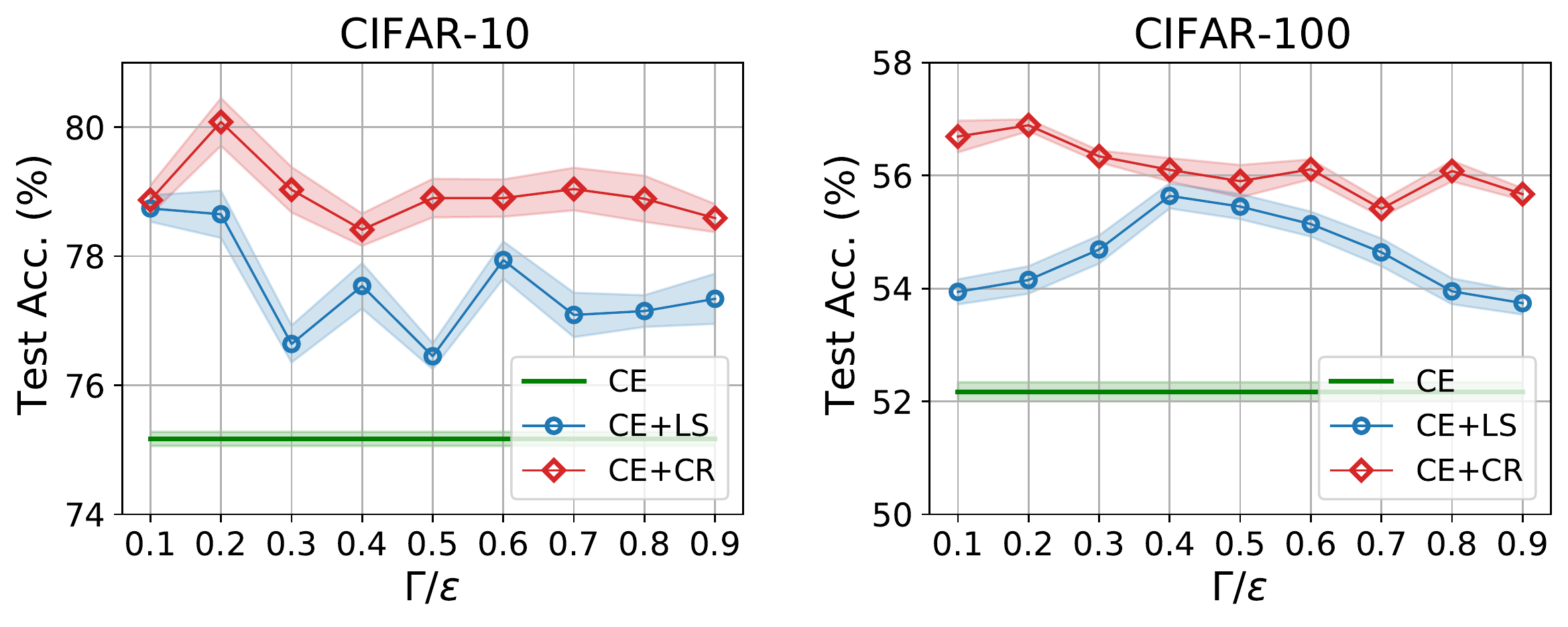} 
		}
		\caption{\small Comparison confidence regularization (CR) with label smooth (LS). $\Gamma$, $\varepsilon$ denote the confidence threshold and smooth coefficient respectively.}
		\label{fig:smooth}
\end{minipage}
}
\end{figure}

\begin{figure}[t]
\centering
\includegraphics[width=12cm]{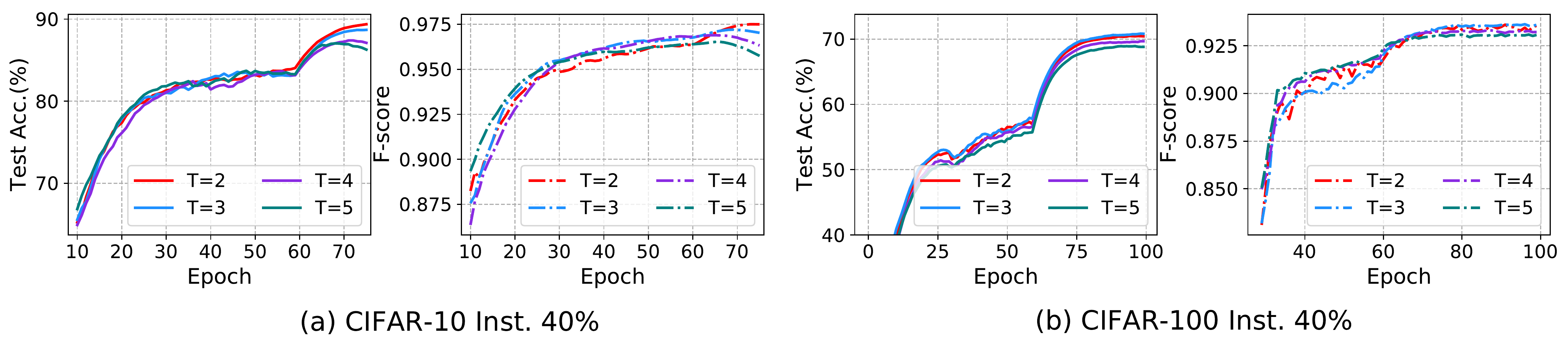} 
\caption{\small The hyper-parameter selection of memory bank size $T$. Test accuracy (\%) and F-score are reported.}
\label{fig:k}
\end{figure}

\subsection{Hyper-parameter selection}
The confidence threshold $\Gamma$ and the size $T$ of the memory bank are two hyper-parameters that need tuning with cross-validation. To study the impact of $T$ and $\Gamma$, we conduct ablation studies and compared the confidence regularization (CR) with label smoothing (LS) as shown in Tab. \ref{tab:ablation_gamma}, Fig. \ref{fig:smooth} and Fig. \ref{fig:k}. 

\noindent\textbf{Confidence threshold $\Gamma$.} We conduct the experiments from two aspects. First, in Tab. \ref{tab:ablation_gamma}, we compare different setting of $\Gamma$ on our framework and the value of $\Gamma$ belongs to $\{0.2, 0.4, 0.6, 0.8\}$. As shown, the different values of $\Gamma$ slightly affect the generalization performance of the model and a relatively small $\Gamma$ is more preferred in our learning framework.

Further, to verify that this confidence regularization can be regarded as a robust loss function in a mild noisy condition, we compare CR with LS on two benchmarks with 40\% symm. label noise. As shown in Fig. \ref{fig:smooth}, our model achieves the best performance when $\Gamma=0.2$ for all benchmarks. For the smooth coefficient $\varepsilon$, the setting is variant for different benchmarks (\textit{e.g.} 0.1 for CIFAR-10, 0.4 for CIFAR-100). Therefore, we recommend $\Gamma=0.2$ for most cases.

\noindent\textbf{Memory bank size $T$.} We conduct experiments with variant settings and plot the testing accuracy and F-score for sample selection in Fig. \ref{fig:k}. Intuitively, the larger size $T$ tends to detect more fluctuation events in the memory bank and further mitigate the selection of boundary examples. As we expected, the classification results and selection accuracy illustrate that our model attains the best performance with a smaller $T$. Therefore, we set $T=3$ for all experiments.\\

\begin{figure}[t]
{
\flushleft
\makeatletter\def\@captype{table}\makeatother
\begin{minipage}{0.45\textwidth}
\flushleft
	\caption{\small Ablation study of each components. The results of \textit{test accuracy} and \textit{F-score} on variant noise labels are reported.}
\scalebox{0.93}{
\begin{tabular}{cl|cc|cc}
\toprule
&  \multicolumn{1}{c|}{\scriptsize \textbf{CIFAR-10}}   & \multicolumn{2}{c|}{\scriptsize Pair. 40\%}      & \multicolumn{2}{c}{\scriptsize Inst. 40\%}  \\ \hline
     &                                              & \scriptsize Acc.  & \scriptsize F-score  & \scriptsize Acc.   & \scriptsize F-score\\\hline
     & \scriptsize \textbf{SFT}    & \scriptsize \textbf{89.74}  & \scriptsize \textbf{0.963}    & \scriptsize \textbf{90.06}   & \scriptsize \textbf{0.969} \\
     & \scriptsize \, w.  Voting                                           & \scriptsize 85.11  & \scriptsize 0.862    & \scriptsize 83.92   & \scriptsize 0.846\\ \hline
     & \scriptsize \, w/o. $\mathcal{R}$                                   & \scriptsize 88.79  & \scriptsize 0.952    & \scriptsize 88.83   & \scriptsize 0.961\\
     & \scriptsize \, w/o. $\mathcal{L}_{\rm {CR}}$                        & \scriptsize 87.14  & \scriptsize 0.944    & \scriptsize 87.36   & \scriptsize 0.948\\
     & \scriptsize \, w/o. $\mathcal{R}$ \& $\mathcal{L}_{\rm {CR}}$       & \scriptsize 86.05  & \scriptsize 0.940    & \scriptsize 86.31   & \scriptsize 0.946\\ 
\bottomrule
\end{tabular} 
}
\label{tab:albation_table}
\end{minipage}
\hspace{1mm}
\makeatletter\def\@captype{figure}\makeatother
\begin{minipage}{0.52\textwidth}
    \centering
	\scalebox{0.75}{
        \includegraphics[width=8.3cm]{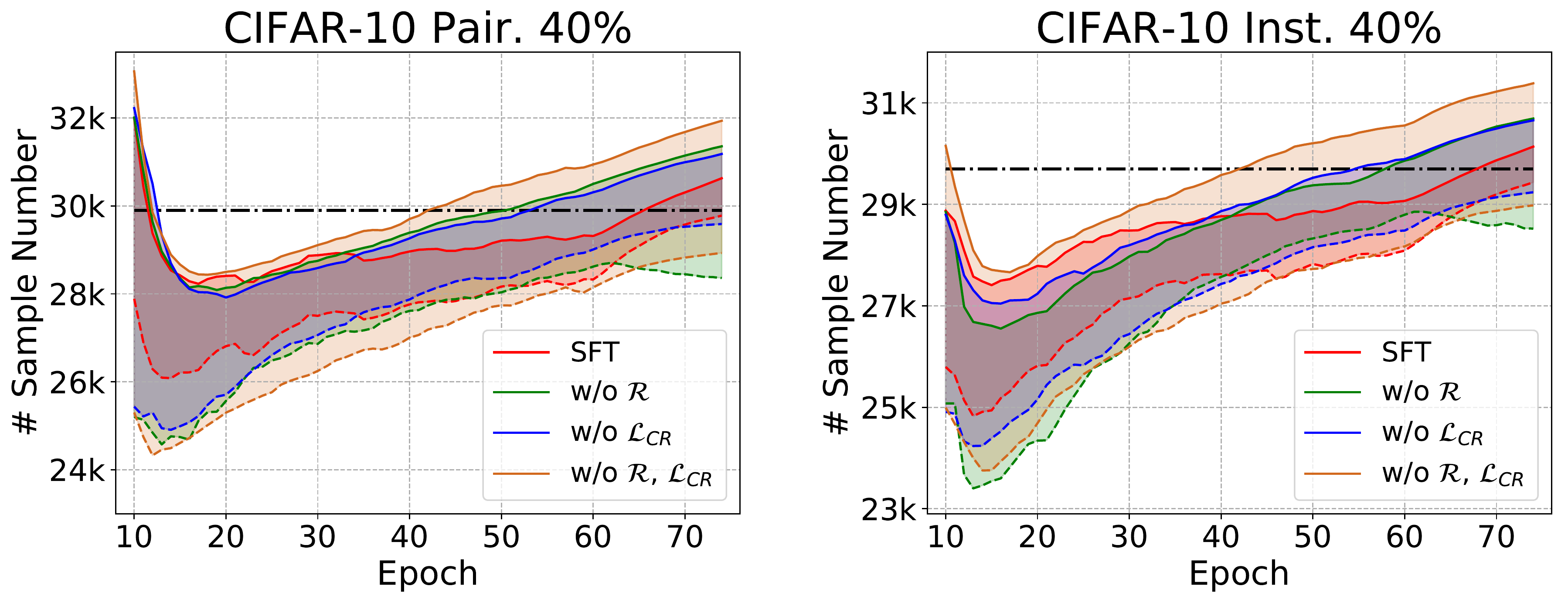} 
		}
		\caption{\small Ablation study of selected samples number. The solid lines and dashed lines denote the number of selected set ${\tilde{\mathcal{D}}}$ and the clean labels in ${\tilde{\mathcal{D}}}$, respectively. The horizontal dashed line denotes the actual clean samples number.}
		\label{fig:albation_fig}
\end{minipage}
}
\end{figure}

\subsection{Ablation study}
\noindent\textbf{Selection criterion.} The majority voting strategy can select samples with high probability of consistent prediction results (right or wrong) in memory bank. Thus, we conduct an ablation study that replaces our criterion with the voting strategy. In Tab. \ref{tab:albation_table}, the significant improvement of test accuracy and F-score compared with voting strategy verifies the superiority of the fluctuation criterion.

\noindent\textbf{Regularization terms.} To directly validate the effectiveness of regularization terms for warming-up and main learning, we remove them in different stages and retrain the model. We evaluate them with the classification accuracy and F-score for the selected results. As shown in Tab. \ref{tab:albation_table}, by removing each component, the performance of the model is degraded. Specifically, test accuracy and F-score averagely decrease \textbf{3.72}\% and \textbf{0.023} on 40\% noise ratio without $\mathcal{R}$ and $\mathcal{L}_{\rm {CR}}$. We also plot the selection curve in Fig. \ref{fig:albation_fig}. With the support of the two terms, the selected subset contains less noisy labels. Meanwhile, red dashed lines in two figures are close to the horizontal dashed line, indicating the boundary samples are almost selected by our framework. 

\section{Conclusion}
In this paper, we propose a simple but effective framework, Self-Filtering, to select clean samples from the noisy training set. We build a memory bank module to store the historical predictions and design the fluctuation criterion for selection. Compared with the small-loss criterion, the fluctuation strategy takes the boundary sample into account and improves the generalization performance for learning with noisy labels. To reduce the accumulated error of the sample selection bias, we propose a confidence-penalization term. By increasing the weight of the misclassified categories with this term, we mitigate the noise effect in learning proceeding and thus the algorithm is robust to label noise. Extensive experiments and studies exhibit the great properties of the proposed framework.

\section{Acknowledgements}
This work was supported in part by National Natural Science Foundation of China (No. 62106129, 62106128, 62176139),  Natural Science Foundation of Shandong Province (No. ZR2021ZD15, ZR2021QF053, ZR2021QF001), Young Elite Scientists Sponsorship Program by CAST (No. 2021QNRC001), and China Postdoctoral Science Foundation (No. 2021TQ0195, 2021M701984).

\bibliographystyle{splncs04}
\bibliography{main.bib}

\clearpage
\end{document}